\documentclass[11pt]{article}

\usepackage[]{acl}

\usepackage{times}
\usepackage{latexsym}

\usepackage[T1]{fontenc}
\usepackage[utf8]{inputenc}

\usepackage{microtype}
\usepackage{graphicx}
\usepackage{enumitem}
\usepackage{tabularx}
\usepackage{multirow}
\usepackage{makecell}
\usepackage{balance}
\usepackage{amsmath,amsfonts,amssymb,amsxtra}
\usepackage{booktabs}

\newcommand{\miniskip}{\vspace*{-.5\baselineskip}}
\newcommand{\shrink}{\vspace*{-.9\baselineskip}}

\newcommand{\bertbase}{BERT$_{\text{\scriptsize{BASE}}}$}
\newcommand{\bertner}{BERT$^{\text{\scriptsize{NER}}}$}
\newcommand{\flairner}{Flair$^{\text{\scriptsize{NER}}}$}
\newcommand{\berttapt}{BERT$_{\text{\scriptsize{TAPT}}}$}
\newcommand{\bertmdtapt}{BERT$^{\text{\scriptsize{MD}}}_{\text{\scriptsize{TAPT}}}$}

\title{Find the Funding: Entity Linking with Incomplete Funding \\
Knowledge Bases}

\author{Gizem Aydin \\
  Radboud University \\
  \texttt{gizemaydin96@gmail.com} \\\And
Seyed Amin Tabatabaei \\
  Elsevier  \\
  \texttt{s.tabatabaei@elsevier.com} \\\AND
  Giorgios Tsatsaronis \\
  Elsevier \\
  \texttt{g.tsatsaronis@elsevier.com}\\\And
  Faegheh Hasibi\\
  Radboud University \\
  \texttt{f.hasibi@cs.ru.nl}}

\begin{document}
\maketitle
\begin{abstract}
Automatic extraction of funding information from academic articles adds significant value to industry and research communities, including tracking research outcomes by funding organizations, profiling researchers and universities based on the received funding, and supporting open access policies.
Two major challenges of identifying and linking funding entities are: (i) sparse graph structure of the Knowledge Base (KB), which makes the commonly used graph-based entity linking approaches suboptimal for the funding domain, (ii) missing entities in KB, which (unlike recent zero-shot approaches) requires marking entity mentions without KB entries as NIL.
We propose an entity linking model that can perform NIL prediction and overcome data scarcity issues in a time and data-efficient manner. Our model builds on a transformer-based mention detection and a bi-encoder model to perform entity linking. We show that our model outperforms strong existing baselines.

\end{abstract}

% !TEX root = ./main.tex
\section{Introduction}

% Text can be obtained from:
% \begin{itemize}
%     \item Introduction chapter, sections 1.1 and 1.2, maybe 1.3
%     \item Related work chapter, section 2.1
% \end{itemize}

Entity Linking (EL) aims to annotate text with corresponding entity identifiers from a Knowledge Base (KB) and is a building block for different tasks, such as document ranking~\cite{Xiong:2017:WED}, entity retrieval ~\cite{Hasibi:2016:EEL}, and question understanding in conversations~\cite{Shang:2021:ERO}. %Reinanda:2015:MRR, Gerritse:2020:GEE, Hasibi:2016:EEL}. %It aims to annotate text with corresponding entity identifiers from a knowledge base (KB)~\cite{balog}.
%A KB is a collection of entities representing uniquely identifiable artefacts in real world. EL often consists of two subtasks: Named Entity Recognition (NER) and Entity Disambiguation (ED). NER corresponds to detecting mentions and their respective types from the text, and ED corresponds to finding which entity in the KB a mention is referring to \cite{balog}.
Recent years have witnessed the flourishing of entity linking approaches for zero-shot~\cite{scalablezeroshot, elq} and open-domain setups~\cite{REL, genre}. %, Zhang:2021:EntQA}.
% especially with the introduction of neural language models~\cite{BERT,elmo}. 
While zero-shot entity linking can generalize to new specialized domains and entity dictionaries, existing approaches cannot perform NIL prediction; i.e., identifying entity mentions without a target entity in a knowledge base and assigning them to NIL. Open-domain entity linkers, on the other hand, build on the availability of rich entity relations and descriptions in KBs. This makes existing EL approaches suboptimal for real-world applications of entity linking in domains with incomplete  knowledge bases, where both in-KB and out-of-KB entities should be identified. %Therefore, domain-specific entity linking and NIL prediction remain as a challenge. 

In this paper, we aim to address entity linking in the funding domain~\cite{GrantExtractor, fundextractdept}, %AckExtract,
which is essential for funding organizations to track the outcome of the research they funded~\citep{ElsPaper} and also helps to comply with open access rules~\citep{GrantExtractor}.
Knowledge bases of funding organizations, either proprietary or  open access (e.g.,  the funding KB Crossref\footnote{\url{https://www.crossref.org}}), contain brief information about entities (e.g., official name and acronym). They also have extremely sparse graph structure with large amount of  missing entities that need to be found from research articles. This implies that EL in the funding domain requires detecting mentions with out-of-KB entities while handling sparse entity relations and descriptions. The approach, should also be able to operate with limited training data, as large public datasets are rarely available for domain-specific applications.

We propose a two-step EL approach, where we first identify entity mentions using task adaptive pre-training~\cite{DontStop} of BERT~\cite{BERT} and then perform Entity Disambiguation (ED) by utilizing a bi-encoder model to learn dense entity and mention representations. 
Our bi-encoder model and training approach using negative sampling are specifically designed to operate with out-of-KB entities. The bi-encoder is followed by a modest feature-based model to map the entities to an entity in KB or NIL. We create two new datasets for EL in the funding domain and compare our mention detection and ED approaches with strong neural and feature-based models. 
We show that our model improves over existing baselines for both entity disambiguation and end-to-end entity linking.

In summary, our contributions include: (i) proposing a data-efficient model for entity linking (with NIL prediction) in funding domain that is efficient and can be used with modest computational power, (ii) improving upon existing EL approaches for funding organization, and (iii) releasing new training and evaluation datasets for entity linking in funding domain.  To our knowledge, this is the first and largest publicly available dataset for entity linking in funding domain. The code and datasets created in this paper are made publicly available.\footnote{\url{https://github.com/informagi/Fund-EL}}

\section{Method}
\begin{figure}[t]
\shrink
    \centering
    \includegraphics[scale=0.4]{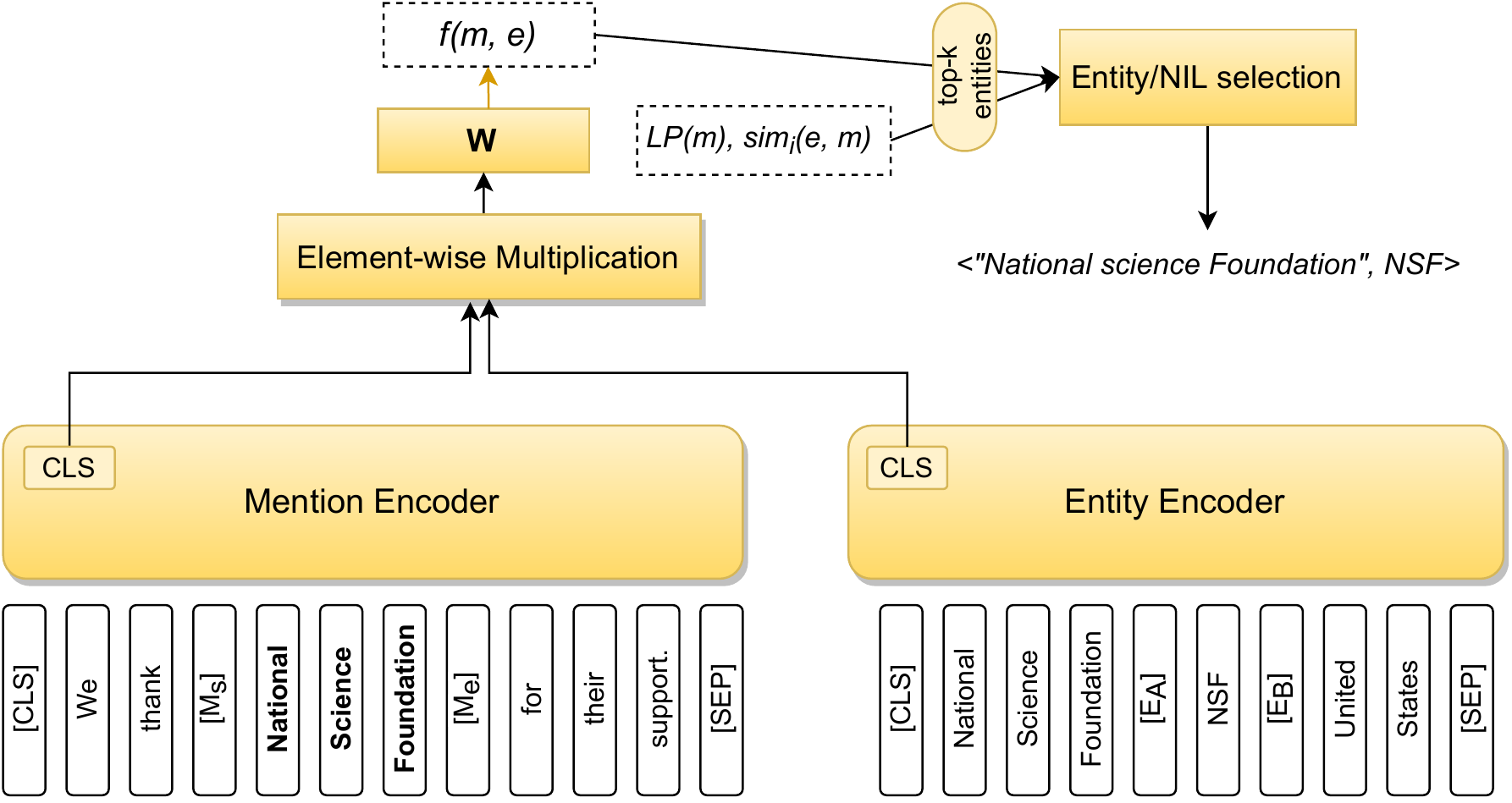}
    \label{fig:my_label}
    \miniskip
    \caption{Overview of our entity disambiguation approach for incomplete proprietary knowledge bases.}
\end{figure}

In this section, we provide a formal definition of the task, followed by the description of our Mention Detection (MD) approach and \underline{F}unding entity \underline{D}isambiguation model, referred to as FunD.
%Our method builds on the two-step entity linking approach \cite{balog}, performing \emph{mention detection} and \emph{entity disambiguation} in two consecutive steps. 

%It is developed with the following design considerations to address the challenges of domain-specific entity liking (and funding domain in particular): 
%(i) it handles incompleteness and sparsity of the knowledge graph by utilizing textual representation of entities,  which otherwise would not be feasible using graph-based approaches commonly used in open-domain entity linking~\cite{REL, graphcitation1, TransE, wikipedia2vec2, mentnorm, additionalref1},
%(ii) it identifies both in-KB and out-of-KB entities, (iii)  it operates with limited domain-specific training data, and 
%(iv) it is computationally efficient and can be used with modest computational power.

\subsection{Task Definition}
We denote $\mathcal{E}$ as the set of entities in a knowledge base, where each entity $e\in \mathcal{E}$ is accompanied by a  textual description. Let $m = (s, t) \in M$ denote  an entity mention with start and end positions $s$ and $t$. Given a document $d = \{w_1, w_2, ..., w_n\}$, our aim is to generate the list:
\begin{equation*}
    \mathcal{L} = \{(\langle s, t\rangle, a) |  1 \leq s \leq t \leq n, a \in \mathcal{E} \cup NIL\},
\end{equation*}
which represents all possible mentions linked to an entity in the KB (\emph{in-KB setup}) or NIL (\emph{out-of-KB setup)}.
This task is similar to zero-shot~\cite{scalablezeroshot} and open domain entity linking~\cite{REL}, but  different from them, entities do not need to have an entry in the KB. 

%provide formal definition of the task with some examples
%Entity Linking (EL) is the task of annotating text with corresponding entity identifiers, from a knowledge base (KB) \cite{balog}. A KB is a collection of entities representing uniquely identifiable artefacts in real world. EL often consists of two subtasks: Named Entity Recognition (NER) and Entity Disambiguation (ED). NER corresponds to detecting \textit{mentions}, strings in text indicating entities, and their respective types. ED corresponds to finding which entity in the KB a mention is referring to \cite{balog}. 

% This work aims to perform EL for annotating articles with organizations that funded the research, and grant numbers. In this context, NER is used to detect mentions of \textit{funding organizations} (ORG) and \textit{grant numbers} (GRT). ED is performed on ORG mentions to determine whether they refer to any funder in the KG, and if so, which one. Formally, the set $T = E_{ORG} \cup M_{GRT}$ is extracted from each input, where:
% \begin{displaymath}
% \begin{split}
% E_{ORG}=\{(m_i,e_i) \mid m_i \in M \land e_i \in \mathcal{E} \land link(m_i)=e_i  \\
% \land \text{ }type(m_i) = ORG\}
% \end{split}
% \end{displaymath}
% and 
% \begin{displaymath}
% M_{GRT} = \{m_i \mid m_i \in M \land type(m_i) = GRT\},
% \end{displaymath}
% in which $M$ and $\mathcal{E}$ denote the set of mentions and the set of entities and the NIL entity.

\subsection{Mention Detection}
For mention detection, we adapt BERT~\cite{BERT} to the funding domain. 
%It is shown that domain-adaptation of BERT improves its performance on downstream tasks in different domains~\cite{DontStop,exBERT,quote3}.
Domain Adaptive Pre-Training (DAPT), while being effective,  requires a large amount of domain-specific text, requires large amount of training data which is not feasible for the funding domain~\cite{Gerritse:2022:ETE, nogueira2019multi}.
We therefore utilize the Task-Adaptive Pre-Training (TAPT)~\cite{DontStop}, which requires a far smaller but more task-relevant training corpus and is proven to be more effective than DAPT.
%Following the TAPT procedure~\cite{DontStop}, 
We train BERT with the Masked Language Model objective on acknowledgments of research papers. We refer to this model as \berttapt. We then fine-tune \berttapt for the mention detection task using IOB tags.
% using unlabelled data related to the task of mention detection in funding domain; i.e., two million acknowledgement sentences of research papers. We refer to this model as \berttapt.

%After the pre-training stage, we fine-tune  \berttapt for the mention detection task. We add a logistic regression layer to the BERT model and classify each token based on IOB (inside, outside, and beginning) tagging. 
%The input to the BERT model is the \texttt{[CLS]} token, followed by the acknowledgement sentence and the \texttt{[SEP]} token. 

\subsection{Entity Disambiguation}
%Our \underline{E}ntity disambiguation model for \underline{P}roprietary \underline{I}ncomplete \underline{K}now\-ledge bases, EPIK, makes use of a dense retrieval approach to find candidate entities. It then applies a classifier to select the best entity or NIL for each mention.
%learn dense entity and mention representation and find candidate entities and then apply a classifier to select the best entity or NIL for each mention.

\paragraph{\textbf{Candidate Entity Selection}}
To obtain the likelihood of an entity being a target link of a mention, we employ a bi-encoder model~\cite{scalablezeroshot} for encoding a mention (with its context) and an entity. 
%Bi-encoder models have been used in a variety of ranking tasks and prove to be effective for zero-shot entity linking~\cite{scalablezeroshot}. Additionally, they provide entity representations based on textual descriptions of entities without requiring graph structure of the knowledge base.  
Our bi-encoder model utilizes two BERT encoders for generating entity and mention representations. The entity encoder takes the structured entity description as the input:
\begin{align*}
    x_e  = \text{BERT}_{\small{\texttt{[CLS]}}} ( &\texttt{[CLS] } \text{val}_{A_1} \texttt{ [E}_A\texttt{] } ... \texttt{ [E}_A\texttt{] }
    \\
    & \text{val}_{A_n} \text{ [E}_B\text{] } \text{val}_B \texttt{ [SEP]}),
\end{align*}
where \texttt{[E}$_A$\texttt{]} and \texttt{[E}$_B$\texttt{]} are two word-piece tokens, selected among the unused tokens of BERT, and val$_{A_i}$  and val$_B$ denote values for entity attributes \emph{A} and \emph{B}. Here \emph{A}  corresponds to names of entities, which is a multi-valued attribute, and \emph{B} is the country of the funding organization.
%While this representation is tailored for funding KB, it can be generalized in a similar vein to other domains.
For the mention encoder, we follow \citet{scalablezeroshot} and obtain mention representations by:
\begin{align*}
    x_m =  \text{BERT}_{\small{\texttt{[CLS]}}}(& \texttt{[CLS] } \text{ctxt}_{left} \texttt{ [M}_s\texttt{]} \text{mention}\\ 
    &  \texttt{[M}_e\texttt{] } 
    \text{ctxt}_{right} \texttt{ [SEP]}),
\end{align*}
where ctxt$_{left}$ and ctxt$_{right}$ represent context words before and after the mention. In this work, both BERT models are initialized with \berttapt. 
The mention-entity score is then obtained by: %to be the probability of the positive class:
\begin{equation}
    f(\textit{m},\textit{e}) = \mathbf{W}^T (x_m \odot x_e),
\end{equation}
where $\odot$ refers to element-wise multiplication of mention representation $x_m$ and entity representation $x_e$, and $\mathbf{W} \in \mathcal{R}^{BERT\times2}$ represents learnable weights. The binary cross-entropy loss $L$ is used to train the model:
\begin{align*}
    L = - \frac{1}{N} \sum_{i=1}^N l_i \log (&f(m_i,e_i)) - (1-l_i)\\
     & \log (1-f(m_i,e_i)),
\label{eq:score}
\end{align*}
where $N$ is the number of training examples and $l_i$ is a binary value that is set to 1 if $e_i$ is the correct entity for mention $m_i$. %Mentions with and without gold entities (NIL annotations) are both used in training, only the latter ones do not have training instances with positive labels.

\paragraph{\textbf{Negative Sampling}}
Following~\citet{googleintern}, we perform training in rounds, where the model obtained in each round is used to produce hard negatives for the next round. Contrary to~\cite{googleintern}, we do not use in-batch random negative sampling, as it provides less diverse random negatives for sparse domain-specific applications compared to open-domain EL.
% do not use the multi-task learning setup with in-batch negative sampling for random negatives, as it provides less diverse random negatives for sparse domain-specific applications compared to open-domain EL. %In other words, in datasets with skewed entity distribution, it is inevitable that the same popular entities are selected as the random negatives for most of the mentions when using in-batch negative sampling.

The following strategy is employed for random and hard negative sampling.
%We identify random and hard negatives using the following strategy. 
In the first round, negative entities of each mention are sampled randomly from the entire KB. 
For the next rounds, both random and hard negatives are used. Hard negatives are entities ranked above the correct entity by the model learned in the previous step. For mentions with out-of-KB entities, the top-K entities are selected as hard negatives (K is set to 10 following~\cite{googleintern}). The number of random negatives for each mention is computed based on the number of hard negatives:
%, computed by. Specifically, the number of random negative samples for each mention is:
%
\begin{equation}
	Neg_r(m) = \lfloor \frac{\sum_{i=1}^{|M|} Neg_h(m)}{|M|} \rfloor,
\end{equation}
where $Neg_r()$ and $Neg_h()$ give number of random and hard negatives, respectively. Using this strategy, we strive a balance between random and hard negatives, while giving hard mentions (i.e., mentions that their correct entities are in low ranks) a larger number of hard negatives.  %This is needed, as such mentions are deemed to be difficult to disambiguate and extra hard negatives enhance optimization. As this strategy strives to balance between random and hard negatives, long tail entities has a better chance to be seen and predicted by the model.% Additionally it enables our model to learn from long tail entities. 

%\paragraph{\textbf{Inference}}
%For inference, we use the pre-computed entity representations and rank entities based on Equation~\eqref{eq:score} for each mention. We then select at most $K'=12$ candidates from the top ranked entities of the bi-encoder's positive class; $K'$ is set empirically based on experiments on the validation set. 

\paragraph{\textbf{Entity or NIL Selection}}
Once we have obtained candidate entities from our bi-encoder model, we turn to mapping each mention to an entity in the knowledge base or NIL. We employ a feature-based model using Gradient Boosting Machine (GBM)~\cite{GBM}. 
%The features are domain-specific and are designed to be lightweight to provide efficient and effective disambiguation. 
Our model utilizes five light-weight features: (i) score obtained by the bi-encoder model, (ii) maximum Levenshtein similarity between the mention and the labels of the candidate entity, (iii) link probability of the mention \cite{balog} obtained by dividing numbers of times a mention appears as a link by total number of occurrences of a term: $P(link|m) = n_{link}(m)/n(m)$, and (iv) commonness score \cite{balog} obtained by $P(e|m) = n(m,e)/\sum_{e'\in \mathcal{E}} n(m, e')$, with $n(m,e)$ denoting number of times that mention $m$ is linked to entity $e$. A mention is linked to the entity with the highest GBM score if higher than threshold $\tau$. %Otherwise, the mention is mapped to NIL.

%During inference, the probability of the positive class is extracted for each mention-candidate entity pair. The mention is linked to the candidate entity with the highest score, if the score is greater than or equal to a threshold, which is selected within the interval [0,1] with using grid search.
\section{Experiments}
\label{sec:exp}

% This section describes our KB (\S\ref{sec:exp:kb}), the datasets we have created for training and evaluation purposes (\S\ref{sec:exp:data}), our experimental setup (\S\ref{sec:exp:setup}), and evaluation metrics (\S\ref{sec:exp:metrics}).

\begin{table}[t]
    \shrink
    \centering
    \begin{tabular}{@{~}lllll@{~}}
    \hline
    \textbf{System}&\textbf{Set}&\textbf{P}&\textbf{R}&\textbf{F1}\\
  \toprule %Xhline{2pt}
    Stanford NER & Test & 73.70 & 75.10  & 74.39 \\
    \flairner &Test  & \textbf{85.83} & 78.02 & 81.74\\
    \bertner &Test  & 79.18 & 86.03&82.46\\
    \bertmdtapt & Test  & 80.28& \textbf{86.54}&\textbf{83.29} \\
    %Flair$^{NER}$ & Validation & \textbf{85.83}&78.02&81.74\\
    \hline
     Stanford NER & Eval &  76.17 & 72.87  & 74.48 \\
     \bertmdtapt & Eval  & \textbf{79.08} & \textbf{85.31} & \textbf{82.08} \\
     \hline
    \end{tabular}
    \caption{Mention detection results on ELFund dataset.}

    \label{tbl:md}
\end{table}

% \begin{table}[t]
%     \centering
%         \caption{Overview of ELFund dataset. The table shows number of articles, number of mention-entity pairs, number of mentions with in-KB and out-of-KB mentions,  and percentage of NIL mentions.}
    
%     % used in training the NER models, number of mention-entity pairs used in training the ED models, and the percentage of NIL mentions among those pairs are shown.}
%     \begin{tabular}{l l l l l l}
%     \hline
%     \textbf{Set}  & \textbf{\#Articles} & \textbf{\#Mentions} & \textbf{\#InKB} & \textbf{\#NIL} & \textbf{NIL\%}  \\
%     \Xhline{2pt}
%     Training  & 29,118 & 95,761 & 77,972& 17,789 & 18.58\% \\
%     Validation  & 991 & 5,618 & 4,749 & 869 & 15.47\%  \\
%     Test  & 3,943 & 19,765 & 16,689& 3,076 & 15.56\%\\
%     Eval & 17,333 & 52,378 & 42,514 & 9,864 & 18.83\%\\
%     \hline
%     \end{tabular}
    
%     \label{tab:goldstats}
% \end{table}

%\paragraph{\textbf{Knowledge Base}}
%
%The knowledge base used for our experiments is a proprietary KB from a major publishing company. The knowledge base contains 25,859 funding organizations, each represented by multiple attributes such as name and country of origin. These entities are sparsely connected to each other, representing affiliation between funding organizations and their hierarchies.

\paragraph{\textbf{Data}}
% \label{sec:exp:data}
We use the Crossref funding registry as our KB, containing information about 25,859 funding organizations.
We create two new datasets for the funding domain, ELFund and EDFund~\cite{edfund},  which are used for experiments.
The datasets are split into training, validation, test, and eval sets, with no overlaps in the training, test, and eval across the two datasets. The validation set is used for searching hyper-parameters.% and monitoring the progress of training. 
The eval set contains completely unseen production data, used for the final evaluation of the models; see Appendix~\ref{sec:data_stats} and \ref{sec:config} for more details about the datasets and experimental setup. %The datasets are made publicly available to the community.

\begin{table*}[t]
    \centering
    \shrink
    \begin{tabular}{ll | ll | lll| lll}
    \hline
    \multirow{2}{*}{\textbf{Method}} & \multirow{2}{*}{\textbf{Set}} 
	& \multicolumn{2}{c |}{All} & \multicolumn{3}{c |}{EE} 
	& \multicolumn{3}{c }{In-KB} \\
	& &
	\multicolumn{1}{l}{\textbf{Acc$_{mic}$}} & \multicolumn{1}{l|}{\textbf{Acc$_{mac}$}} &
	\multicolumn{1}{l}{\textbf{P$_{mic}$}} & \multicolumn{1}{l}{\textbf{R$_{mic}$}} &
	\multicolumn{1}{l|}{\textbf{F1$_{mic}$}} & 
	\multicolumn{1}{l}{\textbf{P$_{mic}$}} & \multicolumn{1}{l}{\textbf{R$_{mic}$}} &
	\multicolumn{1}{l}{\textbf{F1$_{mic}$}} \\

    \hline %{2pt}
    % BI$_{BL}$ (R1) & - & Validation &  60.16	&64.04&50.4&	57.31&	53.63&62.25&60.69	&61.46\\
    % BI$_{BL}$ (R2) & - & Validation & 55.06	&59.13&54.1&	72.96&	62.13&55.31&51.78	&53.49 \\
    % BI$_{AD}$ (R2) & -  & Validation & 86.38&89.45&69.64&79.98&74.45&90&87.56&88.76\\
    Commonness &   Test &83.8&85.81& 53.55& \textbf{88.2}&	66.64 & \textbf{94.22}&82.99	&88.25 \\
    % BI$_{AD}$ & -  & Test & 88.44&90.75&  75.75& 76.27&	76.01&90.8&90.68&	90.74\\
    GBM$_{F26}$ &  Test & 91.02 &\textbf{92.84}  & \textbf{79.11}& 78.67& 78.89 & 93.2 & \textbf{93.29} & 93.25\\
    FunD & Test  &\textbf{91.15} &92.76& 77.44 & 81.14 &\textbf{79.25}& 93.82 & 92.99 &\textbf{93.40}\\
 \hline
    GBM$_{F26}$ & Eval &  90.26&	90.84& \textbf{80.03}&	81.49&	80.75  & 92.69&	\textbf{92.3} &	92.5\\
    FunD & Eval  &\textbf{90.66} & \textbf{91.11} &79.26 &\textbf{85.45}&\textbf{82.24}&\textbf{93.56}  &91.86 &\textbf{92.7} \\
     \hline
    \end{tabular}
        \miniskip
    \caption{Entity disambiguation results on the EDFund dataset. Best results for each set are marked in bold face.}
        \label{tbl:ed}

\end{table*}

\paragraph{\textbf{Evaluation Metrics}}
% \label{sec:exp:metrics}
% Mention all metrics, which one is used for which table. Refer to the original papers. For accuracy, be a bit more extensive and explain the formula.
To evaluate the mention detection step, we use strong matching precision, recall, and F1 score~\cite{conll, gerbil}. The ED and EL tasks are evaluated in three settings: (i) \emph{In-KB}, for mentions linked to entities in KB, (ii) \emph{Emerging Entities (EE)} for Out-of-KB entities; i.e., mentions linked to no entities, and (iii) \emph{All}, for in-KB and emerging entities.  We evaluate the ED task using micro and macro averaged accuracy for  the All setting \cite{NILMentions}. For In-KB and EE settings, we report on micro- and macro- averaged precision, recall, and F1~\cite{gerbil}. %, Cornolti:2013:FBE}.

\section{Results}

%BERT$^{NER}$ slightly outperforms Flair$^{NER}$ in terms of F1 score by an increase of 0.7\%. However, the main difference is the precision and recall values. Flair$^{NER}$ achieves a precision that is 6.6\% higher than that of BERT$^{NER}$, while BERT$^{NER}$ achieves a recall that is 8\% higher. A higher recall is preferred in this study as if a mention is missed completely, there is nothing that can be done about it.

\paragraph{\textbf{Mention Detection}}
Table~\ref{tbl:md} shows the results for the mention detection step. We compare \bertmdtapt with  Stanford NER~\cite{stanfordNER}, \flairner~\cite{Akbik:2018:Flair}, and \bertner~\cite{BERT}. 
%Stanford NER is a feature-based model, used as a robust model by the publishing company, and both \flairner and \bertner models are fine-tuned on the funding domain using ELFund training set. 
The results show that \flairner achieves the highest precision, but the lowest recall compared to the BERT-based models. We also observe that \bertmdtapt outperforms all baselines with respect to the F1 score, showing the importance of task adaptive pre-training when limited data is available.

\begin{table}
    \centering
    \begin{tabular}{l l@{~}l@{~} ll@{~}}
    \hline
    \textbf{MD}  & \textbf{ED}  &\textbf{Setting} &\textbf{F1$_{mic}$} &\textbf{F1$_{mac}$}\\
	\toprule %{2pt}
	Stanford NER & GBM$_{F26}$ & All &	68.43&	69.34 \\
    \bertmdtapt & FunD & All & \textbf{75.81}&\textbf{76.59}\\
    \hline
    Stanford NER & GBM$_{F26}$ & EE & 43.34&71.01 \\
    \bertmdtapt&FunD & EE  &\textbf{52.82} &\textbf{73.68} \\
    \hline
    Stanford NER & GBM$_{F26}$ & In-KB & 77.33& 74.73 \\
    \bertmdtapt&FunD & In-KB
    &\textbf{85.14}&\textbf{81.86} \\
    \hline
    \end{tabular}
        % \miniskip
        \caption{Entity linking results on the ELFund dataset.}

    \label{tab:el}
\end{table}

\paragraph{\textbf{Entity Disambiguation}}
Table~\ref{tbl:ed} presents entity disambiguation results. We compare our ED method, FunD with two baselines: (i) Commonness~\cite{Hasibi:2015:ELQ}, where each mention is linked to the entity with the highest commonness score if it is greater than zero, (ii) GBM$_{F26}$, which is a strong feature-based GBM model with 26 features, ranging from string similarities (e.g. BM25) to statistical features (e.g., commonness and link probability). 
We note that state-of-the-art EL methods, such as REL~\cite{REL}, GENRE~\cite{genre}, and BLINK~\cite{scalablezeroshot} cannot be used as baselines, as they rely on data resources that are not available in our KB and also do not address NIL prediction. We, however, implemented BLINK's bi-encoder with a score threshold, and obtained micro average accuracy of $60.16$, which is a far worse performance compared to other baselines. Table~\ref{tbl:ed} results show that FunD strives a balance between precision and recall and can achieve the best results with respect to F1 in both In-KB and EE setups. This observation is also mirrored with respect to accuracy on the Eval set.

\paragraph{\textbf{Entity Linking}}
Putting the pieces together, we show the results of end-to-end entity linking on test set of ELFund dataset in Table~\ref{tab:el}. We compare our model with the best ED baseline (GBM$_{F26}$) combined with Stanford NER (a fast and strong existing MD model). The results indicate that our MD and ED models improve the existing feature based model by a large margin, reinforcing our previous finding that our models can be successfully applied to the funding domain.

\paragraph{\textbf{Efficiency}}
Finally, we measure the run time of ED model by running it on a random sample of 100 sentences with 306 mentions. We pass 12 candidate entities to both GBM$_{F26}$ and FunD and measure the run time in seconds. The experiment is repeated 10 times on a machine with an Intel Xeon E-2276M (2.80GHz, 32GB RAM) CPU and an NVIDIA Quadro T1000 GPU with 4GB memory. 
Table~\ref{tbl:efficiency} shows that  FunD is four times faster than the feature-based GBM$_{F26}$ model without GPU. The difference is even larger using GPU, as FunD's efficiency is increased, while GBM$_{F26}$ performance does not change with GPU. The inefficiency of the GBM$_{F26}$ model is mostly attributed to the calculation of the hand-crafted features.

\begin{table}
    \centering
    \begin{tabular}{l c c}
    \hline
    \textbf{System}   & \textbf{With GPU} &\textbf{Without GPU}  \\
    \toprule%{2pt}
    % BI$_{AD}$ & $8.83 \pm $ 0.4 & $22.78 \pm 1.56$\\
    FunD & $9.26 \pm 0.47$ &$23.07 \pm 0.78$\\
        GBM$_{F26}$ & 99.2$\pm 0.45$ & 99.2$\pm 0.45$ \\
    \hline
    \end{tabular}
    % \miniskip
    \caption{Efficiency of ED models (in seconds).}
    \label{tbl:efficiency}
\end{table}
\section{Conclusions}
In this paper, we have introduced an entity linking method for funding domain, where the knowledge base has sparse graph structure and limited information is available about entities. The model builds on BERT to perform mention detection, and a bi-encoder model to conduct the entity disambiguation. We compared our method to strong feature-based and zero-shot models and showed that our model can perform NIL assignments and overcome data scarcity issues more efficient and effective than comparable baselines. As future work, we would like to explore the benefit of employing contrastive learning for the highly ambiguous entity mentions, which could provide further robustness to extracting and linking such entities in scientific texts that span across all sciences.

%.
% novel method to address domain-specific entity linking (EL) with incomplete knowledge bases (KBs), and have focused on the application of EL in the research funding domain to demonstrate the value and the benefits of our approach compared to strong feature-based and zero-shot models. Our EL model can perform NIL assignments and overcome data scarcity issues frequently associated with incomplete KBs. The model builds on BERT to perform mention detection, and a bi-encoder model to conduct the entity disambiguation. Experimental evaluation on two benchmarks that have been created for this application, namely EDFund for entity disambiguation, and ELFund for end-to-end entity linking shows that the suggested approach outperforms strong traditional and feature-based models. As future work, we would like to explore the benefit of employing contrastive learning for the highly ambiguous entity mentions, which could provide further robustness to extracting and linking such entities in scientific texts that span across all sciences.
\pagebreak

\section*{Acknowledgements}
We thank Ramadurai Petchiappan, Georgios Cheirmpos, and Efthymios Tsakonas, for their assistance in preparing the final released datasets.
%
%This document has been adapted
%by Steven Bethard, Ryan Cotterell and Rui Yan
%from the instructions for earlier ACL and NAACL proceedings, including those for 
%ACL 2019 by Douwe Kiela and Ivan Vuli\'{c},
%NAACL 2019 by Stephanie Lukin and Alla Roskovskaya, 
%ACL 2018 by Shay Cohen, Kevin Gimpel, and Wei Lu, 
%NAACL 2018 by Margaret Mitchell and Stephanie Lukin,
%Bib\TeX{} suggestions for (NA)ACL 2017/2018 from Jason Eisner,
%ACL 2017 by Dan Gildea and Min-Yen Kan, 
%NAACL 2017 by Margaret Mitchell, 
%ACL 2012 by Maggie Li and Michael White, 
%ACL 2010 by Jing-Shin Chang and Philipp Koehn, 
%ACL 2008 by Johanna D. Moore, Simone Teufel, James Allan, and Sadaoki Furui, 
%ACL 2005 by Hwee Tou Ng and Kemal Oflazer, 
%ACL 2002 by Eugene Charniak and Dekang Lin, 
%and earlier ACL and EACL formats written by several people, including
%John Chen, Henry S. Thompson and Donald Walker.
%Additional elements were taken from the formatting instructions of the \emph{International Joint Conference on Artificial Intelligence} and the \emph{Conference on Computer Vision and Pattern Recognition}.


\bibliography{ref}

\appendix
\section{Dataset Statistics}
\label{sec:data_stats}

\balance

The ELFund and EDFund datasets are created based on scientific articles published before 2017. Expert annotators were asked to identify sentences that contain funding organizations of the research (e.g., X was funded by source Y) and link the organizations to entities in the Crossref KB. ELFund, further contains sentences that could be also automatically identified by a classifier.  Both datasets were annotated by two experts to find the boundary of mentions and their corresponding entities if available. Disagreements were resolved by a third annotator, and mentions with out-of-KB entities are annotated with NIL. We note that ELFund is not a subset of EDFund. 
\begin{table}[h]
    \centering
    \begin{tabular}{@{~}l l@{~}l@{~} l@{~}l@{~}l@{~}@{~}}
    \hline
    \textbf{Set}  & %\#Articles &
    \textbf{\#Articles}$_{\text{\scriptsize{EL}}}$  &
    \textbf{\#Links}$_{\text{\scriptsize{EL}}}$  &
    %  \textbf{\#Articles}$_{\text{\scriptsize{ED}}}$  &
    \textbf{\#Links}$_{\text{\scriptsize{ED}}}$ & \textbf{NIL}$_{\text{\scriptsize{ED}}}$\%  \\
    \Xhline{2pt}
    Train & 22,720 & 67,671   & 95,761 & 18.58\% \\%& 29,118
    Val & 991 &  4,333 & 5,618 & 15.47\%  \\%& 991
    Test & 3,943 &  16,355  & 19,765 & 15.56\%\\%& 3,943
    Eval & 13,851 & 37,495  & 52,378 & 18.83\%\\%& 17,333
    \hline
    \end{tabular}
	\caption{Overview of ELFund and EDFund datasets.For each split, number of articles, number of mentions used in training the NER models, number of mention-entity pairs used in training the ED models, and the percentage of NIL mentions among those pairs are shown.}

    \label{tab:goldstats}
\end{table}

\section{Training Configuration}
\label{sec:config}
%The \berttapt~ model is train the case-preserving version of \bertbase with 2M sentences containing funding information
We train the case-preserving version of \bertbase with 2M sentences containing funding information to obtain the \berttapt~ model. 
%For training of the \berttapt~ model, we first initialize the case-preserving version of \bertbase and train it with 2M sentences containing funding information. 
Unless indicated otherwise, the hyper-parameters recommended by~\citet{DontStop} are used for training.
  The training is done on an NVIDIA Tesla K80 GPU with 12 GB of memory with a batch size of 2048 through gradient accumulation and for one epoch (1000 steps). 
We further fine-tune \berttapt for the mention detection task on the ELFund dataset. The fine-tuning process is done for 3 epochs with batch size of 8. We refer to this model as \bertmdtapt. 
For disambiguation, the bi-encoder model is trained on the EDFund dataset with a learning rate of $2\times10^{-5}$ and batch size of 16. The training is performed in 4 rounds, each round consisting of 2 epochs. In the first round, 3 random negatives are used for each mention. The score threshold $\tau$ is set to 0.042 using grid search. Following~\cite{scalablezeroshot}, the mention and entity representations are limited to 64 and 256 tokens, respectively. 

%\label{sec:appendix}
%
%This is an appendix.
\end{document}